\documentclass{article}
\usepackage{spconf,amsmath,graphicx}
\ninept

\usepackage{amsfonts}
\usepackage{dsfont}
\usepackage{siunitx}
\usepackage{epstopdf}
  \usepackage[caption=false,font=footnotesize]{subfig}

\DeclareMathOperator{\Tr}{Tr}
\newcommand{\B}{\mathbf{B}} 
\newcommand{\Z}{\mathbf{Z}} 
\newcommand{\Y}{\mathbf{Y}} 
\newcommand{\X}{\mathbf{X}} 
\newcommand{\E}{\mathbf{E}} 
\newcommand{\M}{\mathbf{M}} 
\newcommand{\D}{\mathbf{D}} 
\newcommand{\R}{\mathbf{R}} 
\newcommand{\N}{\mathbf{N}} 
\newcommand{\V}{\mathbf{V}} 

\title{Hyperspectral image superresolution: An edge-preserving convex formulation}
\name{
\begin{tabular}{c}
Miguel Sim\~{o}es$^{1,2}$, Jos\'{e} Bioucas-Dias$^{1}$, Luis B. Almeida$^{1}$, Jocelyn Chanussot$^{2,3}$
\thanks{This work was partially supported by the Funda\c{c}\~{a}o para a Ci\^{e}ncia e Tecnologia, Portuguese Ministry of Science and Higher Education, projects PEst-OE/EEI/0008/2013,  PTDC/EEI-PRO/1470/2012, and grant SFRH/BD/87693/2012.}
\end{tabular}
}
\address{$^1$ Instituto de Telecomunica\c{c}\~{o}es, Instituto
Superior T\'{e}cnico, Universidade de Lisboa, Portugal.\\
$^2$GIPSA-Lab, Grenoble Institute of Technology, France.\\
$^3$Faculty of Electrical and Computer Engineering, University of Iceland.}

\begin{document}
%
\maketitle
\begin{abstract}
Hyperspectral remote sensing images (HSIs) are characterized by having a low spatial resolution and a high spectral resolution, whereas multispectral images (MSIs) are characterized by low spectral and high spatial resolutions.  These complementary characteristics have stimulated active research in the inference of images with high spatial and spectral resolutions from HSI-MSI pairs.

In this paper, we formulate this data fusion problem as the minimization of a convex objective function containing two data-fitting terms and an edge-preserving regularizer. The data-fitting terms are quadratic and account for blur, different spatial resolutions, and additive noise; the regularizer, a form of vector Total Variation, promotes aligned discontinuities across the reconstructed hyperspectral bands.

The optimization described above is rather hard, owing to its non-diagonalizable linear operators, to the non-quadratic and non-smooth nature of the regularizer, and to the very large size of the image to be inferred. We tackle these difficulties by tailoring the Split Augmented Lagrangian Shrinkage Algorithm (SALSA)---an instance of the Alternating Direction Method of Multipliers (ADMM)---to this optimization problem.  By using a convenient variable splitting and by exploiting the fact that HSIs generally ``live'' in a low-dimensional subspace, we obtain an effective algorithm that yields state-of-the-art results, as illustrated by experiments.

\end{abstract}
\section{Introduction}
\label{sec:intro}
A spectral image, or \emph{data cube}, is a set of 2D images, also termed \emph{bands}, representing the reflectance or radiance of a scene in different parts of the electromagnetic spectrum.
In the remote sensing field, it is common to distinguish between hyperspectral and multispectral images (HSIs and MSIs, respectively). The difference is application-dependent, but HSIs typically have high spectral resolution in the visible, near-infrared, and shortwave infrared spectral range~\cite{Bioucas-Dias2012}. As a result of this high resolution, HSIs  have a large number of two-dimensional bands (\textit{i.e.}, images), each one corresponding to a narrow band of the EM spectrum.
On the other hand, MSIs generally offer a higher spatial resolution, but each band covers a larger range of the spectrum, resulting in a much smaller total number of them.

It is of interest to fuse the information from these two data sources, to obtain images with high spectral and spatial resolutions. A related data fusion problem that has been extensively studied is pansharpening, which is the fusion of multispectral and panchromatic images, the latter being single-band images usually covering the visible and near-infrared spectral ranges~\cite{Amro2011}.

Recently, some techniques dedicated to the fusion of HSIs and MSIs have been proposed. A recent trend is to associate this problem with the linear spectral unmixing one.
Introduced in~\cite{Charles2011} for HSIs but with older works exploring similar ideas for MSIs~\cite{Zurita2008}, this approach consists in learning a dictionary from the
HSI and then using it to reconstruct the MSI via sparse regression. The estimate of the original high resolution HSI is then obtained from the regression coefficients and from the dictionary.  A similar approach was taken in~\cite{Kawakami2011a, Huang2013}, in which the dictionary-based representation was interpreted as a linear mixing model.
Alternatively, one can unmix both images, and try to find a correspondence between them \cite{Yokoya2012, Song2013}.
In \cite{Hardie2004, Zhang2009, Zhang2012, Wei2013}, a fully Bayesian approach was followed, by imposing prior distributions on the problem.
A different but related problem was studied in~\cite{Khan2009}, where the HSIs were fused, not with several bands, but with one panchromatic band.



The remainder of this work is organized as follows. Section~\ref{sec:method} describes the data fusion method, including the proposed model and the formulation of the optimization problem. 
It is followed by Section~\ref{sec:exp}, which presents some experimental results, and by Section~\ref{sec:conclusions}, which concludes this work.


\section{Data Fusion Method}
\label{sec:method}

\subsection{Observation Model}
\vspace{-0.2cm}
For notational convenience, the representation followed in this work will consider HSIs and MSIs to be two-dimensional matrices, where each line corresponds to a spectral band, containing the lexicographically ordered pixels of that band.
Let $\Y_h \in \mathbb{R}^{L_h \times n_h}$ denote the observed hyperspectral data with $L_h$ bands and spatial dimension $n_h$,   $\Y_m \in \mathbb{R}^{L_m \times n_m}$ denote  the observed multispectral data, which have $L_m<L_h$ bands and spatial dimension $n_m>n_h$, and $\Z \in \mathbb{R}^{L_h \times n_m}$ denote the high spatial and spectral resolution data to be estimated.

With these definitions in place, the hyperspectral measurements are modeled as
\begin{equation} \label{eq:model_y_h}
\Y_h = \Z \B \M + \N_h,
\end{equation}
where $\B \in \mathbb{R}^{n_m \times n_m}$ is the matrix representation of the sensor point spread function in the spatial resolution of $\Z$, assumed to be band-independent, with circular boundary conditions.
Matrix $\M \in \mathbb{R}^{n_m \times n_h}$ accounts for a subsampling of the image $\Z$. We assume uniform subsampling defined on a subset of the spatial grid associated with the bands of $\Z$. Therefore, the columns of $\M$ are a subset of the columns of the identity matrix. $\N_h$ represents i.i.d. noise.

The multispectral measurements are modeled as
\begin{equation} \label{eq:model_y_m}
\Y_m = \R \Z + \N_m,
\end{equation}
where $\R \in \mathbb{R}^{L_m \times L_h}$ holds in its rows the spectral responses of the  multispectral instrument, one per band, and $\N_m$ represents i.i.d. noise.

Matrices $\B$ and $\R$ can be built by taking into account the registration between multispectral and hyperspectral images, and information about the sensors provided by the manufacturer of the instruments. These matrices can, however, also be estimated from the data, by formulating two coupled least squares problems. We don't provide details here due to lack of space, but the experimental results presented in Section \ref{sec:exp} use this form of estimation, instead of relying on externally supplied information about the sensors.

\subsection{Dimensionality reduction}
\label{sec:dimreduction}
\vspace{-0.2cm}
Hyperspectral data are highly correlated: the spectral vectors, of size $L_h$, normally ``live'' in a subspace of dimension much lower than $L_h$~\cite{Bioucas-Dias2012}. Therefore, we can write
\begin{equation} \label{eq:Z}
\Z = \E \X,
\end{equation}
where $\E \in \mathbb{R}^{L_h \times s}$ is a matrix whose $s$ columns span the same subspace as the columns of $\Z$, and $\X \in \mathbb{R}^{s \times n_m}$ are the representation coefficients.
Small values of $s$, i.e., $s \ll L_h$, translate into a description of the data in a relatively low-dimensional space.
This representation reduces the number of variables to be estimated, yielding a much faster and more accurate estimation than if we worked in the original space of $\Z$.

Matrix $\E$ can be estimated in two different ways. One of them consists in interpreting $\E$ as the unmixing matrix from a linear unmixing problem. We can estimate this matrix using, for example, the Vertex Component Analysis algorithm (VCA)~\cite{Nascimento2005}. The other way consists in building $\E$ from the left singular vectors of $\Y_h$ that correspond to the $s$ largest singular values. If $\N_h=\mathbf{0}$ and all discarded singular values are zero, this representation spans the true signal subspace. If the former condition on $\N_h$ is not obeyed but $\N_h$ is i.i.d., the maximum likelihood estimate of the subspace is still given by the first $s$ singular vectors of $\Y_h$. However, if the noise is non-i.i.d., the estimation of the subspace is more complex. See, for example, \cite{Bioucas-Dias2008} for details, and for algorithms oriented to subspace estimation in hyperspectral applications.


\subsection{Regularization}
\vspace{-0.2cm}
The problem that we are trying to solve is strongly ill-posed, and therefore needs adequate regularization. The regularizer that we use is given by
\begin{equation}
\varphi(\X) \; {\buildrel\rm def\over=} \; \sum_{j=1}^{n_m} \sqrt{\sum_{i=1}^{s} \Big\{\big[(\X \D_h)_{ij}\big]^2 + \big[(\X \D_v)_{ij}\big]^2\Big\}}
\end{equation}
where $(.)_{ij}$ denotes the element in the $i$th row and $j$th column of a matrix, and the products by matrices $\D_h$ and $\D_v$ compute the horizontal and vertical discrete differences of an image, respectively, with periodic boundary conditions. This regularizer is a form of vector Total Variation (VTV)~\cite{Bresson2008}. Its purpose is to impose sparsity in the absolute gradient distribution of an image, meaning that transitions between adjacent pixels of an image should be smooth, except for a small number of them, which should coincide with details such as edges. This vector form of the regularizer promotes solutions in which edges and other details are aligned among the different bands, which is not the case for the non-vector form. This regularizer has previously been used in a pansharpening application~\cite{He2012} and in the denoising of hyperspectral images~\cite{Yuan2012}. We apply it to the reduced-dimensionality data $\X$, and not to $\Z$ itself. One may raise the question of whether it makes sense to apply the regularizer to $\X$. This is in fact so, since the subspace spanned by $\E$ is the same as the one where $\Z$ resides (or an approximation), and by regularizing $\X$ we are indirectly regularizing $\Z$.

\begin{figure*}[!t]
\centering
\subfloat[Observed HSI (false color, crop).]{\includegraphics[scale=1.2, trim=0 0 0 70, clip=true]{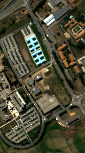}%
\label{fig:paviahs}}
\hfil
\subfloat[Observed panchromatic image (crop).]{\includegraphics[scale=.4, trim=0 0 0 208, clip=true]{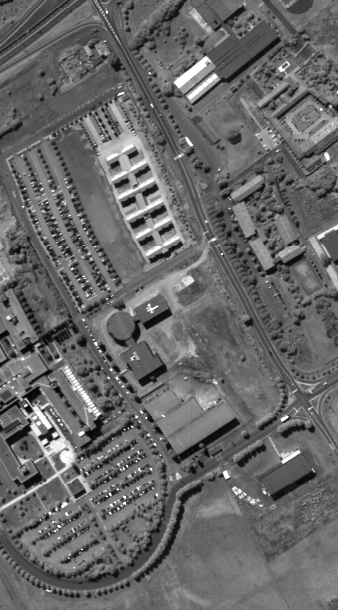}%
\label{fig:paviapan}}
\hfil
\subfloat[Proposed method's result (false color, crop).]{\includegraphics[scale=.3, trim=0 0 0 277, clip=true]{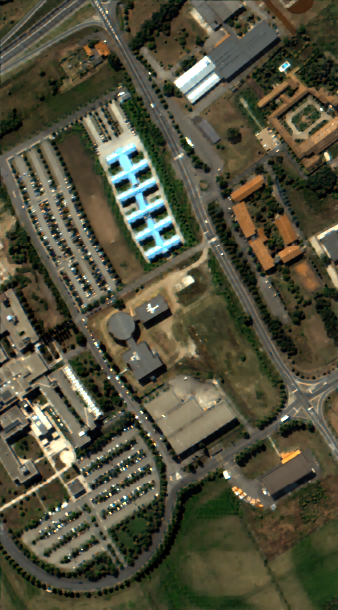}%
\label{fig:pavia50hat}}
\hfil
\subfloat[BT's result (false color, crop).]{\includegraphics[scale=.3, trim=0 0 0 277, clip=true]{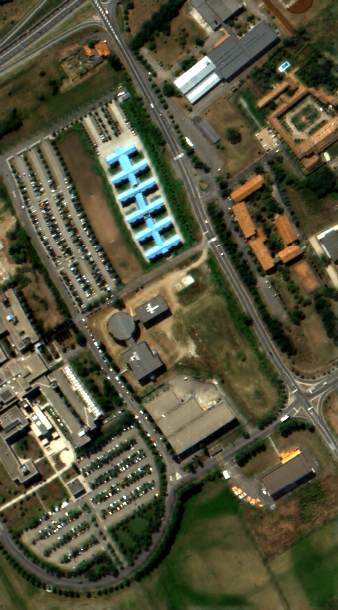}%
\label{fig:pavia50bhatBT}}
\hfil
\caption{Hyperspectral + panchromatic fusion on the Pavia dataset.}
\label{fig:pavia}
\vspace{-0.2cm}
\end{figure*}

\subsection{Optimization problem}
\vspace{-0.2cm}
Let $\| \X \|_F \; {\buildrel\rm def\over=} \; \sqrt{\Tr(\X \X^T)}$ be the Frobenius norm of $\X$. We can now formulate an optimization problem based on our model with the proposed regularizer:
\begin{equation} \label{eq:optimizationproblem}
\begin{aligned}
& \underset{\X}{\text{minimize}}
& & \frac{1}{2}\|\Y_h - \E\X\B\M \|_F^2 + \frac{\lambda_{m}}{2}\|\Y_m - \R\E\X \|_F^2 \\
& & & \quad + \lambda_{\varphi} \varphi (\X \D_h, \X \D_v).\\	
\end{aligned}
\end{equation}
The first two terms account for the data misfit and the last term is the regularizer. The parameters $\lambda_m$ and $\lambda_{\varphi}$ control the relative significances of the various terms.

Problem \eqref{eq:optimizationproblem} is convex, but is rather hard to solve, due to the nature of the regularizer, which is non-quadratic and non-smooth. Additional difficulties are raised by the large size of $\X$ (the variable to be estimated) and by the presence of the downsampling operator $\M$ in one of the quadratic terms, preventing a direct use of the Fourier transform in optimizations involving this term. We deal with these difficulties by using the Split Augmented Lagrangian Shrinkage Algorithm (SALSA)~\cite{Afonso2011}, which is an instance of the Alternating Direction Method of Multipliers (ADMM) optimization method. This allows us to decouple the minimization of $\X$ into a series of much simpler problems, by means of the variable splitting technique, in which four auxiliary variables, $\V_1$ to $\V_4$, are used:
\begin{equation} \label{eq:constraint}
\begin{aligned}
& \underset{\X}{\text{minimize}}
& & \frac{1}{2}\|\Y_h - \E\V_1\M \|_F^2 + \frac{\lambda_{m}}{2}\|\Y_m - \R\E\V_2 \|_F^2 \\
& & & \quad + \lambda_{\varphi} \varphi(\V_3, \V_4)\\
& \text{subject to}
& & \X \B = \V_1,\\
& & & \X = \V_2,\\
& & & \X\D_h = \V_3,\\
& & & \X\D_v = \V_4.		
\end{aligned}
\end{equation}

SALSA is an algorithm which solves a complex optimization problem through an iteration on a set of much simpler problems. The constraints are taken into account by minimizing the augmented Lagrangian of the problem. The minimization with respect to $\X$ is a quadratic problem with a block cyclic system matrix, which can be efficiently solved by using the Fast Fourier Transform (FFT) algorithm. Minimizing with respect to the auxiliary variables is done by solving three different problems whose solutions correspond to three Moreau proximity operators~\cite{Combettes2006}. The minimization with respect to $\V_1$ is a quadratic problem which is efficiently solved via FFTs, and the minimization relative to $\V_2$ is also quadratic; these two problems involve a matrix inverse which can be computed in advance. Finally, the minimization with respect to $\V_3$ and $\V_4$ corresponds to a pixel-wise \textit{vector soft-thresholding} operation. In~\cite{Afonso2011}, conditions for SALSA convergence were established, which are satisfied in this case. Note that an alternative approach to solve this problem consists in employing a primal-dual method~\cite{Chambolle2011, Condat2013}. Unlike our approach, these methods do not require the use of auxiliary variables. However, SALSA has the flexibility of being able to deal with an arbitrary number of terms in the optimization function and, since it yields diagonalizable subproblems, it is much faster in problems of this kind, according to our experience.

\begin{figure}[t]

\begin{minipage}[b]{1.0\linewidth}
  \centering
  \centerline{\includegraphics[scale=.3]{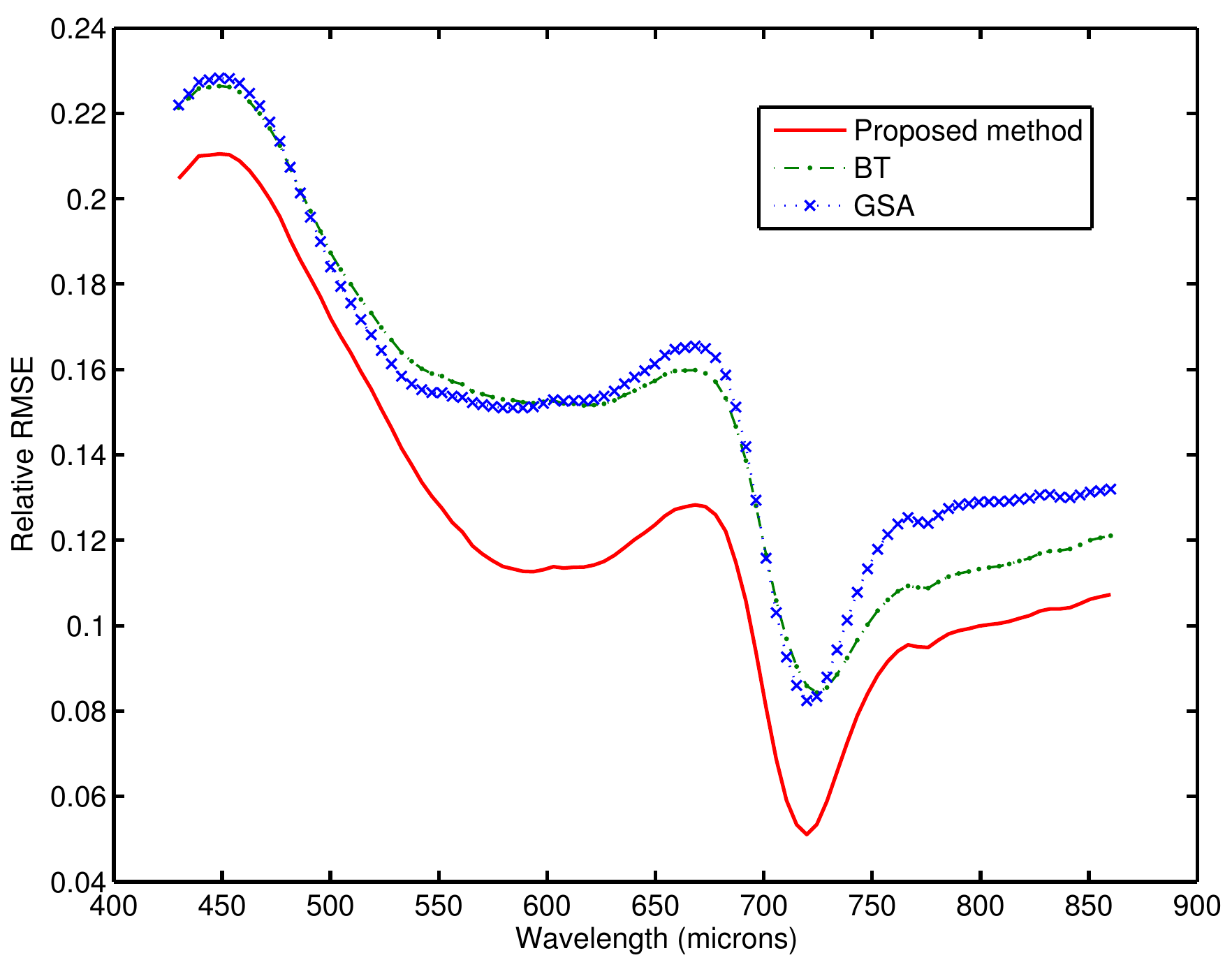}}
\end{minipage}
\vspace{-.4cm}
\caption{Relative Root-mean-square error (RMSE) between the estimated image and the ground truth, for the different bands (for the three best methods).}
\label{fig:spectrum}
\vspace{-.4cm}
\end{figure}

\section{Experimental Results} \label{sec:exp}
\vspace{-0.2cm}
Since the pansharpening literature is, generally speaking, much more established than the literature on the fusion of hyper- and multispectral data, our method was first compared against a number of published pansharpening methods in the fusion of a hyperspectral image with a panchromatic one (the situation in which $L_m=1$). Subsequently, it was compared with a published HSI-MSI fusion method.

Two datasets were used to test the different algorithms, one partially synthetic and the other one using only real-life images.
The first dataset was based on a standard hyperspectral image (Pavia University, see Fig.~\ref{fig:pavia}). This image, acquired with the Reflective Optics System Imaging Spectrometer (ROSIS)~\cite{Kramer1994}, was used to synthesize HSIs and MSIs according to (\ref{eq:model_y_h}) and (\ref{eq:model_y_m}), respectively, and was also used as ground truth.
To synthesize the lower spectral resolution MSI images, the spectral response of the IKONOS satellite was used. This satellite captures one panchromatic and four multispectral bands~\cite{Kramer1994}. The panchromatic band's response was used for the tests made in Subsection \ref{sec:exp_B} and the multispectral bands' response for those made in Subsection \ref{sec:exp_C}. For synthesizing the HSIs, we performed blurring using the Stark-Murtagh filter~\cite{Starck1994} followed by subsampling with factor $1/4$ in both the horizontal and vertical directions.
The scattered light in hyperspectral applications is split among many more bands than in multispectral applications, usually yielding a lower SNR per band in the former case than in the latter one. To account for this, we added Gaussian noise with an SNR of 30 dB to the HSI and with an SNR of 40 dB to the MSI.

\begin{figure*}[!t]
\centering
\subfloat[Observed HSI (false color, crop)]{\includegraphics[scale=1.8, trim=0 0 0 35, clip=true]{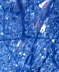}%
\label{fig:parishs}}
\hfil
\subfloat[Proposed method's result (false color, crop)]{\includegraphics[scale=.6, trim=0 0 0 105, clip=true]{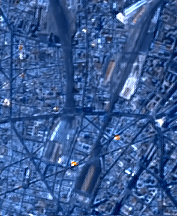}%
\label{fig:parishat}}
\hfil
\subfloat[GSA's result (false color, crop)]{\includegraphics[scale=.6, trim=0 0 0 105, clip=true]{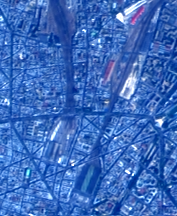}%
\label{fig:parishat_GSA}}
\caption{Hyperspectral + panchromatic fusion on the Paris dataset.}
\label{fig:paris}
\vspace{-0.5cm}
\end{figure*}

The second dataset, taken above Paris and shown on Fig.~\ref{fig:parishs}, was acquired by two instruments on board the Earth Observing-1 Mission (EO-1) satellite:
the Hyperion instrument, generating hyperspectral images at a spatial resolution of 30 meters, and the Advanced Land Imager (ALI),
which provides both multispectral and panchromatic images at resolutions of 30 and 10 meters, respectively~\cite{Middleton2003}. Since the MSI and the HSI have the same spatial resolution, we only fused the panchromatic image with the hyperspectral one (see Section~\ref{sec:exp_B}).

To quantitatively evaluate the quality of the results on the Pavia dataset, for which a ground truth image was available,  we used three indices proposed in the literature: the \textit{Erreur Relative Globale Adimensionnelle de Synth\`{e}se} (ERGAS), proposed in~\cite{Wald2000},
the Spectral Angle Mapper (SAM),
and an index based on the Universal Image Quality Index (UIQI), proposed by Wang \textit{et al.}~\cite{Wang2002}
(the implementation used in this work considers a sliding window of $32 \times 32$ pixels).
When working on real-life images we had no access to the ground-truth, and used an index proposed by Alparone \textit{et al.} for these conditions, the so-called Quality with No Reference (QNR)~\cite{Alparone2008}, which is based on an evaluation of the spatial and spectral distortions of the estimated image, $D_s$ and $D_\lambda$, respectively.

In the implementation of our algorithm, we estimated matrix $\E$ using VCA for both datasets. Estimating the subspace with truncated SVD and keeping the ten singular vectors corresponding to the ten largest singular values allows us preserve at least $99.95\%$ of the energy of the original image. Since the subspace estimated by VCA shares the dimension of the subspace estimated by SVD~\cite{Nascimento2005}, we made $s=10$. Note that, due to the random nature of VCA, the results presented correspond to the average of ten runs of our algorithm. To choose the values of the algorithm's parameters, we first found the optimal values for each situation, and computed the corresponding quality indices. We then chose a set of parameter values that were the same for all situations but yielded quality indices that were very close to the previously found optimal ones. These parameter values were $\lambda_m = 1$ and $\lambda_{\varphi}=10^{-2}$ when fusing a HSI with a panchromatic image and $\lambda_m = 1$ and $\lambda_{\varphi}=5 \times 10^{-4}$ when fusing a HSI with a MSI. Additionally, we performed two pre-processing steps on the hyperspectral data: first, uncalibrated or very noisy bands were removed; afterward, the data were denoised by estimating a reduced rank subspace of $\Y_h$ by means of an SVD with $s=10$, and then projecting the data onto this subspace.

In~\cite{Boyd2011}, a stopping criterion was proposed for problems solved via ADMM. We verified that this criterion worked well, always yielding less than 200 iterations. Given this, we ran the algorithm for the fixed number of 200 iterations in every case.

\subsection{Fusion of hyperspectral and panchromatic images}  \label{sec:exp_B}
\vspace{-0.2cm}
A number of algorithms drawn from the pansharpening literature were used for comparison with our method:
the Gram-Schmidt method (GS) and its adaptive version (GSA)~\cite{Aiazzi2007}, the Fast Intensity-Hue-Saturation Fusion Technique (FIHS)~\cite{Tu2004}, the Principal Component Analysis method (PCA)~\cite{Tu2004}, the Brovey Transform fusion method (BT)~\cite{Tu2004}, and the Box High-pass Filtering method (HPF)~\cite{Amro2011}.
%
The results for the Pavia dataset can be seen in Table~\ref{tab:pavia_hs_pan} and Fig.~\ref{fig:pavia}. A comparative analysis of the errors along the different bands is shown in Fig.~\ref{fig:spectrum}.
%
%
%
The results for the Paris dataset can be seen on Table~\ref{tab:paris_hs_pan} and Fig.~\ref{fig:paris}. 
We found that the published pansharpening methods seem to not deal well with the fact that the panchromatic image's spectral range does not overlap with a large number of hyperspectral channels. 

\subsection{Fusion of hyperspectral and multispectral images}   \label{sec:exp_C}
\vspace{-0.2cm}
For these tests we used the Pavia dataset, and compared to Zhang \textit{et al.}'s method (ZM)~\cite{Zhang2009}. This method does not estimate the spatial blur, so we estimated it in a manner similar to the way it is estimated by our method. ZM assumes that the input HSIs and MSIs are represented with the same spatial resolution. Since this is not the case in the problem we are addressing, we upsampled the HSIs to the resolution of the MSIs for input to ZM. Following the authors' lead, we chose the decomposition level of the Nondecimated Wavelet Transform to be three. Additionally, due to input restrictions of the implementation of ZM that was available, we only worked on a part of the image (a section of the image with $200 \times 200$ pixels corresponding to the bottom left corner). In an implementation of the algorithm developed with MATLAB\textsuperscript{\textregistered} 7.13, running on a desktop PC equipped with an Intel\textsuperscript{\textregistered} Xeon\textsuperscript{\textregistered} CPU (at 3.20GHz) and 16 GB of RAM memory, the proposed method took approximately 35 seconds to perform the fusion, on average.
%
%
%
%
The results can be seen in Table~\ref{tab:pavia_hs_ms}. 

\begin{table}[!t]\footnotesize
\renewcommand{\arraystretch}{1.3}
\caption{Pavia dataset: results for hyperspectral + panchromatic fusion.}
\label{tab:pavia_hs_pan}
\centering
\begin{tabular}{l||c|c|c}
\multicolumn{1}{r}{} & ERGAS & SAM & UIQI\\
\hline
\hline
Proposed & \textbf{3.813} & 4.856 & \textbf{0.937} \\ 
\hline
GS & 4.960 & 5.494 & 0.885 \\ 
\hline 
GSA & 4.587 & 5.116 & 0.893 \\ 
\hline 
FIHS & 4.813 & 5.255 & 0.895 \\ 
\hline 
PCA & 7.609 & 9.448 & 0.755 \\ 
\hline 
BT & 4.533 & \textbf{4.550} & 0.919 \\ 
\hline 
HPF & 5.573 & 6.151 & 0.866 \\ 
\hline
\hline
\end{tabular}
\end{table}
\begin{table}[!t]\footnotesize
\renewcommand{\arraystretch}{1.3}
\caption{Paris dataset: results for hyperspectral + panchromatic fusion.}
\label{tab:paris_hs_pan}
\centering
\begin{tabular}{l||c|c|c}
\multicolumn{1}{r}{} & $\text{D}_{s}$ & $\text{D}_{\lambda}$ & $\text{QNR}$ \\
\hline
\hline
Proposed & 0.024 & 0.108 & \textbf{0.871}\\
\hline
GS & 0.092 & 0.104 & 0.814\\
\hline
GSA & \textbf{0.022} & 0.156 & 0.826\\
\hline
FIHS & 0.125 & 0.136 & 0.756\\
\hline
PCA & 0.105 & \textbf{0.089} & 0.816\\
\hline
BT & 0.159 & 0.144 & 0.720\\
\hline
HPF & 0.061 & 0.122 & 0.824\\
\hline
\hline
\end{tabular}
\vspace{-0.5cm}
\end{table}
\begin{table}[!t]\footnotesize
\renewcommand{\arraystretch}{1.3}
\caption{Pavia dataset (bottom left corner): results for hyperspectral + multispectral fusion.}
\label{tab:pavia_hs_ms}
\centering
\begin{tabular}{l||c|c|c}
\multicolumn{1}{r}{} & ERGAS & SAM & UIQI\\
\hline
\hline
Proposed & \textbf{1.213} & \textbf{1.956} & \textbf{0.995}\\
\hline
Zhang & 5.919 & 4.375 & 0.881\\
\hline
\hline
\end{tabular}
\vspace{-0.5cm}
\end{table}
%
%

\section{Conclusions} \label{sec:conclusions}
\vspace{-0.2cm}
We presented a flexible method to perform the fusion of hyperspectral data with either panchromatic or multispectral images, in order to obtain data with high resolution both in the spectral and in the spatial domain. This fusion problem is closely related to the pansharpening one, but presents new challenges: hyperspectral images are much larger than the multispectral images normally used in pansharpening, the different sources of data do not always spectrally overlap, and when fusing with multispectral images, the high spatial resolution data have several bands, instead of a single one.

The fusion problem was formulated as a convex optimization one, and was solved via the Split Augmented Lagrangian Shrinkage Algorithm (SALSA), an instance of the Alternating Direction Method of Multipliers (ADMM). By using a convenient variable splitting and by exploiting the fact that HSIs generally ``live'' in a low-dimensional subspace, we obtained an effective algorithm which compares quite favorably to several published methods, both on simulated and on real-life datasets. In the fusion process, the method can estimate both the relative spectral response and the spatial response of the sensors from the data.

\section{Acknowledgement}
\vspace{-0.2cm}
We gratefully acknowledge Prof. Paolo Gamba for providing the ROSIS Pavia
University data set. We also gratefully acknowledge Dr. Yifan Zhang
for providing the source code for~\cite{Zhang2009} and also Dr. Giorgio Licciardi and Dr.
Gemine Vivone for providing data and source code used in the comparisons.

%



\bibliographystyle{IEEEbib}
\newpage
\bibliography{refs}

\begin{thebibliography}{10}

\bibitem{Bioucas-Dias2012}
J.~Bioucas-Dias, A.~Plaza, N.~Dobigeon, M.~Parente, Q.~Du, P.~Gader, and
  J.~Chanussot,
\newblock ``Hyperspectral unmixing overview: Geometrical, statistical, and
  sparse regression-based approaches,''
\newblock {\em IEEE J. Sel. Topics Appl. Earth Observ.}, vol. 5, no. 2, pp.
  354--379, 2012.

\bibitem{Amro2011}
I.~Amro, J.~Mateos, M.~Vega, R.~Molina, and A.~Katsaggelos,
\newblock ``A survey of classical methods and new trends in pansharpening of
  multispectral images,''
\newblock {\em EURASIP J. Adv. Signal Process.}, vol. 2011, no. 1, pp. 79--100,
  2011.

\bibitem{Charles2011}
A.~Charles, B.~Olshausen, and C.~Rozell,
\newblock ``Learning sparse codes for hyperspectral imagery,''
\newblock {\em IEEE J. Sel. Topics Signal Process.}, vol. 5, no. 5, pp.
  963--978, 2011.

\bibitem{Zurita2008}
R.~Zurita-Milla, J.~Clevers, and M.~Schaepman,
\newblock ``Unmixing-based {L}andsat {TM} and {MERIS FR} data fusion,''
\newblock {\em IEEE Geosci. Remote Sens. Lett.}, vol. 5, no. 3, pp. 453--457,
  2008.

\bibitem{Kawakami2011a}
R.~Kawakami, J.~Wright, Y.~Tai, Y.~Matsushita, M.~Ben-Ezra, and K.~Ikeuchi,
\newblock ``High-resolution hyperspectral imaging via matrix factorization,''
\newblock in {\em Proc. IEEE Computer Society Conf. Computer Vision},
  Providence, RI, June 2011, pp. 2329--2336.

\bibitem{Huang2013}
B.~Huang, H.~Song, H.~Cui, J.~Peng, and Z.~Xu,
\newblock ``Spatial and spectral image fusion using sparse matrix
  factorization,''
\newblock {\em IEEE Trans. Geo. Remo. Sens.}, vol. 52, no. 3, pp. 1693--1704,
  2013.

\bibitem{Yokoya2012}
N.~Yokoya, T.~Yairi, and A.~Iwasaki,
\newblock ``Coupled nonnegative matrix factorization unmixing for hyperspectral
  and multispectral data fusion,''
\newblock {\em IEEE Trans. Geosci. Remote Sens.}, vol. 50, no. 2, pp. 528--537,
  2012.

\bibitem{Song2013}
H.~Song, B.~Huang, K.~Zhang, and H.~Zhang,
\newblock ``Spatio-spectral fusion of satellite images based on dictionary-pair
  learning,''
\newblock in {\em Informat. Fusion}. 2013, http://
  dx.doi.org/10.1016/j.inffus.2013.08.005.

\bibitem{Hardie2004}
R.~Hardie, M.~Eismann, and G.~Wilson,
\newblock ``{MAP} estimation for hyperspectral image resolution enhancement
  using an auxiliary sensor,''
\newblock {\em IEEE Trans. Image Process.}, vol. 13, no. 9, pp. 1174--1184,
  2004.

\bibitem{Zhang2009}
Y.~Zhang, S.~{De Backer}, and P.~Scheunders,
\newblock ``Noise-resistant wavelet-based bayesian fusion of multispectral and
  hyperspectral images,''
\newblock {\em IEEE Trans. Geosci. Remote Sens.}, vol. 47, no. 11, pp.
  3834--3843, 2009.

\bibitem{Zhang2012}
Y.~Zhang, A.~Duijster, and P.~Scheunders,
\newblock ``A bayesian restoration approach for hyperspectral images,''
\newblock {\em IEEE Trans. Geosci. Remote Sens.}, vol. 50, no. 9, pp.
  3453--3462, 2012.

\bibitem{Wei2013}
Q.~Wei, N.~Dobigeon, and J.~Tourneret,
\newblock ``Bayesian fusion of multi-band images,''
\newblock in {\em arXiv [cs.CV]}. 2013, http://arxiv.org/abs/1307.5996.

\bibitem{Khan2009}
M.~Khan, J.~Chanussot, and L.~Alparone,
\newblock ``Pansharpening of hyperspectral images using spatial distortion
  optimization,''
\newblock in {\em Proc. IEEE Int. Conf. Image Processing}, Cairo, 2009, pp.
  2853--2856.

\bibitem{Nascimento2005}
J.~Nascimento and J.~Bioucas-Dias,
\newblock ``Vertex component analysis: A fast algorithm to unmix hyperspectral
  data,''
\newblock {\em IEEE Trans. Geosci. Remote Sens.}, vol. 43, no. 4, pp. 898--910,
  2005.

\bibitem{Bioucas-Dias2008}
J.~Bioucas-Dias and J.~Nascimento,
\newblock ``Hyperspectral subspace identification,''
\newblock {\em IEEE Trans. Geosci. Remote Sens.}, vol. 46, no. 8, pp.
  2435--2445, 2008.

\bibitem{Bresson2008}
X.~Bresson and T.~Chan,
\newblock ``Fast dual minimization of the vectorial total variation norm and
  applications to color image processing,''
\newblock {\em Inverse Probl. and Imag.}, vol. 2, no. 4, pp. 455--484, 2008.

\bibitem{He2012}
X.~He, L.~Condat, J.~Chanussot, and J.~Xia,
\newblock ``Pansharpening using total variation regularization,''
\newblock in {\em IEEE Int. Geosci. and Remote Sens. Symp.}, Munich, 2012, pp.
  166--169.

\bibitem{Yuan2012}
Q.~Yuan, L.~Zhang, and H.~Shen,
\newblock ``Hyperspectral image denoising employing a spectral--spatial
  adaptive total variation model,''
\newblock {\em IEEE Trans. Geosci. Remote Sens.}, vol. 50, no. 10, pp.
  3660--3677, 2012.

\bibitem{Afonso2011}
M.~Afonso, J.~Bioucas-Dias, and M.~Figueiredo,
\newblock ``An augmented {L}agrangian approach to the constrained optimization
  formulation of imaging inverse problems.,''
\newblock {\em IEEE Trans. Image Process.}, vol. 20, no. 3, pp. 681--95, 2011.

\bibitem{Combettes2006}
P.~Combettes and V.~Wajs,
\newblock ``Signal recovery by proximal forward-backward splitting,''
\newblock {\em Multiscale Model. Sim.}, vol. 4, no. 4, pp. 1168--1200, 2005.

\bibitem{Chambolle2011}
A.~Chambolle and T.~Pock,
\newblock ``A first-order primal-dual algorithm for convex problems with
  applications to imaging,''
\newblock {\em J. Math. Imaging Vis.}, vol. 40, no. 1, pp. 120--145, 2011.

\bibitem{Condat2013}
L.~Condat,
\newblock ``A primal-dual splitting method for convex optimization involving
  {L}ipschitzian, proximable and linear composite terms,''
\newblock {\em J. Optimiz. Theory Appl.}, vol. 158, no. 2, pp. 460--479, 2013.

\bibitem{Kramer1994}
H.~Kramer,
\newblock {\em Observation of the Earth and its Environment},
\newblock Springer, Berlin, 1994.

\bibitem{Starck1994}
J.~Starck and F.~Murtagh,
\newblock ``Image restoration with noise suppression using the wavelet
  transform,''
\newblock {\em Astron. Astrophys.}, vol. 288, pp. 342--348, 1994.

\bibitem{Middleton2003}
E.~Middleton, S.~Ungar, D.~Mandl, L.~Ong, S.~Frye, P.~Campbell, D.~Landis,
  J.~Young, and N.~Pollack,
\newblock ``The {Earth Observing One (EO-1)} satellite mission: over a decade
  in space,''
\newblock {\em IEEE J. Sel. Topics Appl. Earth Observ.}, vol. 6, no. 2, pp.
  243--256, 2013.

\bibitem{Wald2000}
L.~Wald,
\newblock ``Quality of high resolution synthesised images: Is there a simple
  criterion?,''
\newblock in {\em Proc. 3rd Conf. Fusion of Earth Data: Merging Point
  Measurements, Raster Maps and Remotely Sensed Images}, Sophia Antipolis,
  2000, pp. 99--103.

\bibitem{Wang2002}
Z.~Wang and A.~Bovik,
\newblock ``A universal image quality index,''
\newblock {\em IEEE Signal Process. Lett.}, vol. 9, no. 3, pp. 81--84, 2002.

\bibitem{Alparone2008}
L.~Alparone, B.~Aiazzi, S.~Baronti, A.~Garzelli, F.~Nencini, and M.~Selva,
\newblock ``Multispectral and panchromatic data fusion assessment without
  reference,''
\newblock {\em Photogramm. Eng. Remote Sens.}, vol. 74, no. 2, pp. 193--200,
  2008.

\bibitem{Boyd2011}
S.~Boyd, N.~Parikh, E.~Chu, B.~Peleato, and J.~Eckstein,
\newblock ``Distributed optimization and statistical learning via the
  alternating direction method of multipliers,''
\newblock {\em Found. Trends Mach. Learn.}, vol. 3, no. 1, pp. 1--122, 2011.

\bibitem{Aiazzi2007}
B.~Aiazzi, S.~Baronti, and M.~Selva,
\newblock ``Improving component substitution pansharpening through multivariate
  regression of {MS+Pan} data,''
\newblock {\em IEEE Trans. Geosci. Remote Sens.}, vol. 45, no. 10, pp.
  3230--3239, 2007.

\bibitem{Tu2004}
T.~Tu, P.~Huang, C.~Hung, and C.~Chang,
\newblock ``A fast intensity hue saturation fusion technique with spectral
  adjustment for {IKONOS} imagery,''
\newblock {\em IEEE Geosci. Remote Sens. Lett.}, vol. 1, no. 4, pp. 309--312,
  2004.

\end{thebibliography}

\end{document}